\documentclass[final]{cvpr}

\usepackage{times}

\usepackage{comment}

\usepackage{graphicx}
\usepackage{subcaption}
\usepackage{float}

\usepackage{booktabs}                   
\usepackage{multirow}
\usepackage{array}
\usepackage{paralist}
\usepackage{enumitem}

\usepackage{graphicx}                  
\usepackage{mathtools}
\usepackage{amssymb,amsmath}   

\usepackage[pagebackref=true,breaklinks=True,colorlinks=true,bookmarks=false,citecolor=blue,linkcolor=blue]{hyperref}

\usepackage{pifont}
\usepackage{xspace}
\usepackage{diagbox}
\usepackage{xcolor}
\usepackage[export]{adjustbox}
\usepackage{sidecap}

\def\etal{et~al.}			  
\def\eg{e.g.,~}               
\def\ie{i.e.,~}               

\newcommand{\xmark}{\text{\ding{55}}}
\newcommand{\cmark}{\text{\ding{51}}}
\newcommand{\Paragraph}[1]
{\vspace{1mm} \noindent\textbf{#1}}

\newcommand{\secref}[1]{Section~\ref{sec:#1}}
\newcommand{\figref}[1]{Figure~\ref{fig:#1}} 
\newcommand{\tabref}[1]{Table~\ref{tab:#1}}
\newcommand{\eqnref}[1]{Eq.~\ref{eq:#1}}

\long\def\ignorethis#1{}
\newcommand {\yb}[1]{{\color{red}\textbf{}#1}\normalfont}

\newcommand {\ybc}[1]{{\color{blue}\textbf{Yan-Bo:}#1}\normalfont}

\newlength\paramargin
\newlength\figmargin
\newlength\subfigmargin
\newlength\secmargin
\newlength\subsecmargin
\newlength\tabmargin
\newlength\eqmargin

\def\cvprPaperID{2230} 
\def\confName{ICCV}
\def\confYear{2021}

\newcommand{\mycomment}[1]{}

\newcommand{\hy}[1]{\textcolor{blue}{\textbf{hy}: #1}}
\begin{document}

\title{Unsupervised Sound Localization via Iterative Contrastive Learning}
 

\author{
Yan-Bo Lin$^{1}$\quad
Hung-Yu Tseng$^{2}$\quad
Hsin-Ying Lee$^{3}$\quad
Yen-Yu Lin $^{1}$\quad
Ming-Hsuan Yang $^{2,4}$
\\
$^{1}$National Yang Ming Chiao Tung University\quad
$^{2}$University of California, Merced\quad \\
$^{3}$Snap Inc. \quad
$^{4}$Google Research\\
}

\maketitle
\begin{abstract}

Sound localization aims to find the source of the audio signal in the visual scene.
However, it is labor-intensive to annotate the correlations between the signals sampled from the audio and visual modalities, thus making it difficult to supervise the learning of a machine for this task.
In this work, we propose an iterative contrastive learning framework that requires no data annotations.
At each iteration, the proposed method takes the 1) localization results in images predicted in the previous iteration, and 2) semantic relationships inferred from the audio signals as the pseudo-labels. 
We then use the pseudo-labels to learn the correlation between the visual and audio signals sampled from the same video (intra-frame sampling) as well as the association between those extracted across videos (inter-frame relation).
Our iterative strategy gradually encourages the localization of the sounding objects and reduces the correlation between the non-sounding regions and the reference audio. 
Quantitative and qualitative experimental results demonstrate that the proposed framework performs favorably against existing unsupervised and weakly-supervised methods on the sound localization task.

\end{abstract}  
\vspace{-4mm}
\section{Introduction}\label{sec:intro}
\begin{figure}[t!]
    \centering
	\includegraphics[width=0.9\linewidth]{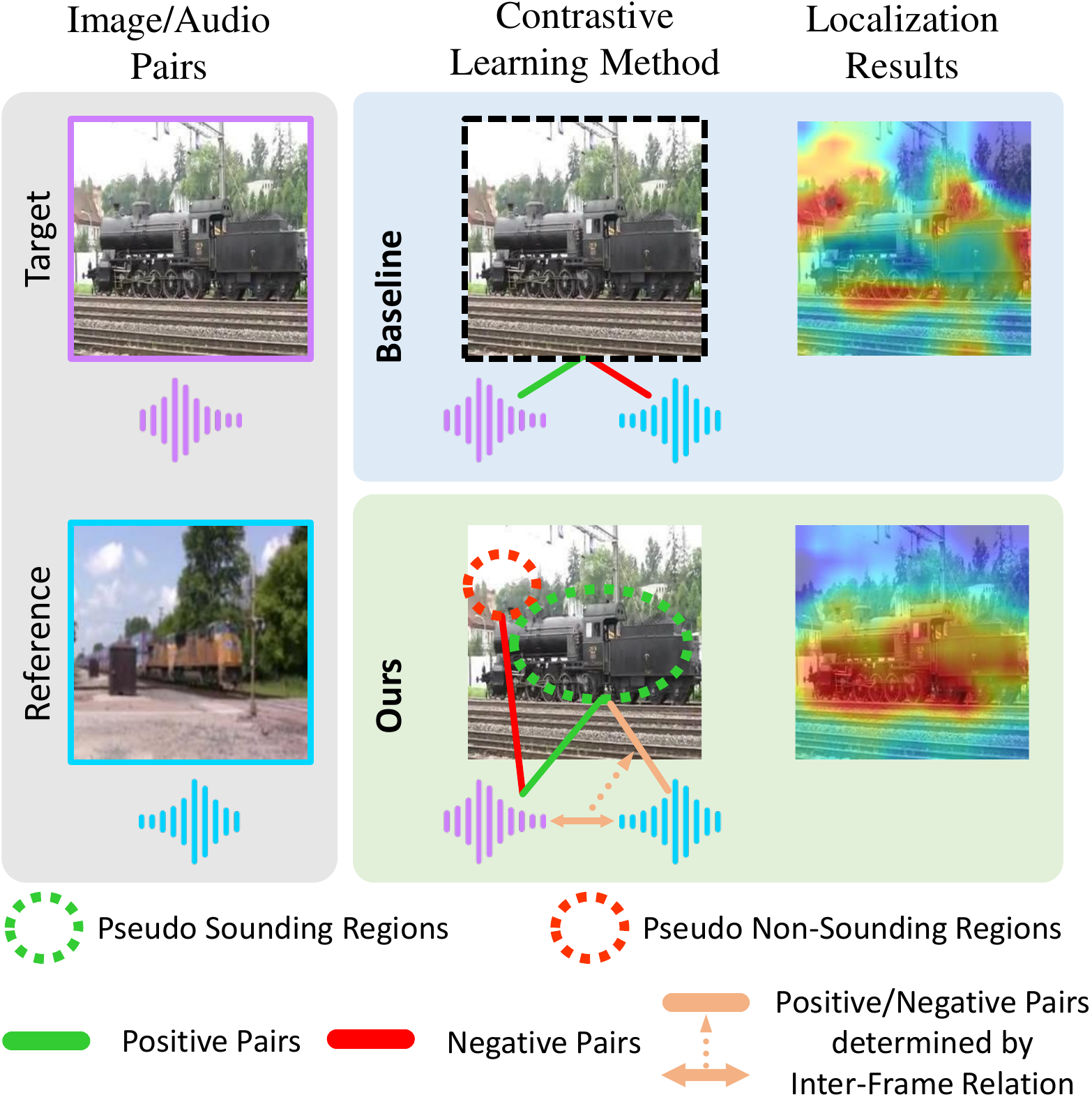}
	\vspace{-2mm}
    \caption{
    \textbf{Unsupervised sound localization via iterative contrastive learning.} 
    (\textit{Baseline}) Existing contrastive learning usually takes the image-audio pairs sampled from the same video frame as the positive pairs, and those extracted from different videos as the negative pairs. 
    (\textit{Ours}) 
    %
    %
    The proposed iterative approach exploits the \textbf{intra-frame sampling} that takes the sounding and non-sounding regions predicted in the previous training epoch as the \emph{pseudo}-labels (green and red dashed circles), and the \textbf{inter-frame relation} that provides additional positive or negative correlations between the image and audio sampled across different videos where the correlations are determined by observing the relationships in the audio modality (positive in this example).
    }
	\label{fig:teaser}
	\vspace{-3.5mm}
\end{figure}

\vspace{\secmargin}
Multisensory signals (\eg vision, hearing, and touching) provide rich information for human beings to perceive the surrounding environments.
These cues from different modalities are usually closely related and thus enable human beings to perform complicated tasks in our daily lives.
Take vision and audio as an example, one can easily imagine a lightning scene upon hearing thunders, associate multiple objects with their sources on a noisy street, and identify and converse with friends in a crowded cocktail party. 
In this work, we target the \emph{sound localization} task~\cite{av_nips20_loc,av_eccv20_mms_loc, av_cvpr18_lls,av_tpami20_lls} that aims to identify the sounding region in the image, as the example shown in \figref{teaser}.
Sound localization is an emerging research topic since it is the nexus of various audio-visual applications such as audio-visual source separation~\cite{av_cvpr20_sep-gesture,av_eccv18_sep,co_sep_iccv19,av_iccv19_mpnet,sofm_iccv19,pix,av_iclr21_AudioScope,av_cvpr21_co_learning,av_cvpr21_gao2021VisualVoice,av_cvpr20_PreviewAudio} and audio-visual event localization/parsing/recognition~\cite{av_eccv20_avvp,eccv18_avel,av_iccv19_DAM,AVSDN,av_cvpr21_av_parsing,my_accv20_av-trans,av_iclr21_lee2021crossattentional}.

Sound localization methods based on supervised learning entail a large amount of training data with the annotated sound-visual associations.
%
Although Senocak~\etal~\cite{av_cvpr18_lls,av_tpami20_lls} collect $5000$ audio-image pairs from the Flickr-Sound database~\cite{av_nips16_soundnet} with bounding box annotations of the sounding regions, the amount of labeled data is not sufficient to train a deep learning model in a fully-supervised fashion.
%
%
%
%
%
Moreover, it is challenging to scale up the efforts to collect a large labeled dataset since the annotators need to meticulously observe visual and audio signals simultaneously.

Semi-supervised~\cite{av_cvpr18_lls,av_tpami20_lls}, weakly-supervised~\cite{av_eccv20_mms_loc}, and self-supervised learning frameworks~\cite{av_cvpr19_deep_cluster,av_arxiv_curricumlum_av_clutser,av_nips20_loc} are proposed to overcome the limited data issue.
The weakly-supervised methods~\cite{av_eccv20_mms_loc} require audio-visual event labels, and existing self-supervised methods rely on a pre-defined number of clusters~\cite{av_cvpr19_deep_cluster,av_arxiv_curricumlum_av_clutser} or require videos of single sounding sources~\cite{av_nips20_loc}.
\mycomment{
\yb{
Particularly, self-supervised approaches are designed to leverage audio-visual clusters~\cite{av_cvpr19_deep_cluster,av_arxiv_curricumlum_av_clutser}, or sounding object dictionaries~\cite{av_nips20_loc}.
However, such weakly-supervised and self-supervised frameworks rely on prior knowledge such as the pre-defined number of clusters~\cite{av_cvpr19_deep_cluster,av_arxiv_curricumlum_av_clutser}, audio-visual event labels~\cite{av_eccv20_mms_loc}, or single-source videos for building up sounding object dictionaries~\cite{av_nips20_loc}.
}
\hy{why mention 4,5 here? and what is the drawback of 18? and it suggests that we are not self-supervised}
\ybc{revised}
}
Furthermore, the semi-supervised methods~\cite{av_cvpr18_lls,av_tpami20_lls} using audio-visual correspondences alone as the supervision is less effective since a scene may contain non-sounding or ambient regions, which leads to the association between the incorrect sounding regions and reference audio signals.
These issues hamper the performance of sound localization in unconstrained scenarios where the numbers of sound sources are usually unknown and there may exist objects unseen during training.

In this work, we propose an iterative contrastive representation learning algorithm that does not require any prior assumption or labels for the sound localization task.
Starting from conventional contrastive learning~\cite{av_cvpr18_lls,av_tpami20_lls}, we use the sound localization model obtained in the \emph{previous} epoch to estimate the sounding and non-sounding regions as the pseudo-labels for the current epoch.
With such pseudo regions, the model is encouraged to disassociate non-sounding or ambient regions from object sounds and thus explores more negative training samples for contrastive learning.
In addition to the relationships between the audio and visual signals within an instance, we correlate audio signals \emph{across} instances.
For instance, if the audio clips of two different instances are semantically similar, the image and audio \emph{across} the two instances should be positively correlated and can then serve as a positive pair for contrastive learning, and vice versa.
We show an example of two train sounds across instances in \emph{inter-frame relation} of \figref{teaser}.
Such a strategy alleviates typical contrastive learning methods from differentiating the representations of the related sounding object and audio signals across instances, and provides more reliable guidance to learn a sound localization model.
%

We evaluate the proposed method on the Flickr-Sound ~\cite{av_cvpr18_lls,av_tpami20_lls} and the MUSIC-Synthetic~\cite{av_nips20_loc} datasets using the consensus intersection over union (cIoU) and area under curve (AUC) as evaluation metrics.
Both qualitative and quantitative results demonstrate the effectiveness of the proposed method on the sound localization task.
The main contributions of this work are summarized as follows:
\begin{compactitem}
  \item We propose an iterative contrastive learning algorithm to tackle the sound localization task without any data annotations.
  \item Our method not only leverages regions of interests, but also exploits non-sounding regions as well as the relationship across audio instances to jointly learn the audio and visual representations.
  \item Qualitative and quantitative experimental results on the benchmark dataset demonstrate that the proposed method performs favorably against the state-of-the-art weakly supervised and unsupervised approaches. 
\end{compactitem}

\section{Related Work}
\vspace{\secmargin}
\Paragraph{Self-Supervised Audio-Visual Representation Learning.} 
Inherent correlation among different modalities of a video provides supervisory signals for learning a deep neural network model.
Information sources used in existing self-supervised audio-visual representation learning methods can be broadly categorized as follows. 
First, audio-visual pairs are extracted from a video clip as positive association. 
The assumption is that the audio and visual features extracted from the same video clip should be strongly correlated ~\cite{av_iccv17_look,av_eccv18_obj_that_sound,av_nips16_soundnet,av_eccv16_abSound,avt_nips20_VersatileNet,av_nips20_xdc,av_nips20_CrossLabelling,av_iclr21_activeContrastive}. 
In addition, these schemes differentiate the features extracted from unpaired video clips.
Furthermore, some concurrent methods~\cite{av_cvpr21_agreementAVID,av_cvpr21_RAVID} jointly consider the correlations within each modality or across different modalities (i.e., audio and vision).
Different from~\cite{av_cvpr21_agreementAVID,av_cvpr21_RAVID} that learn visual information of an entire image, our method leverages pseudo-annotations to provide training guidance from both sounding and non-sounding regions.
Second, video temporal information~\cite{av_eccv18_Owens,av_nips18_coop} is explored to determine strong or weak correlation. 
%
Given a video sequence, a few methods sample the audio and visual features from the same time frame as strong correlation and consider those across different frames as weak correlation for the representation.
Third, spatial relations among image regions are exploited. 
Since the binaural recording techniques (spatial audio) preserve the spatial information of the sound origins, some approaches~\cite{25d,av_nips20_spatial_alignment,360gen,av_cvpr20_tell_left_right,av_eccv20_sep-stereo,my_aaai21_avconsistency,av_cvpr21_withoutBinaural,lu2019self} jointly model the visual and audio information spatially to construct spatial audio generation systems or learn representations for downstream tasks.
%


\begin{figure*}[t!]
    \centering
	\includegraphics[width=0.9\linewidth]{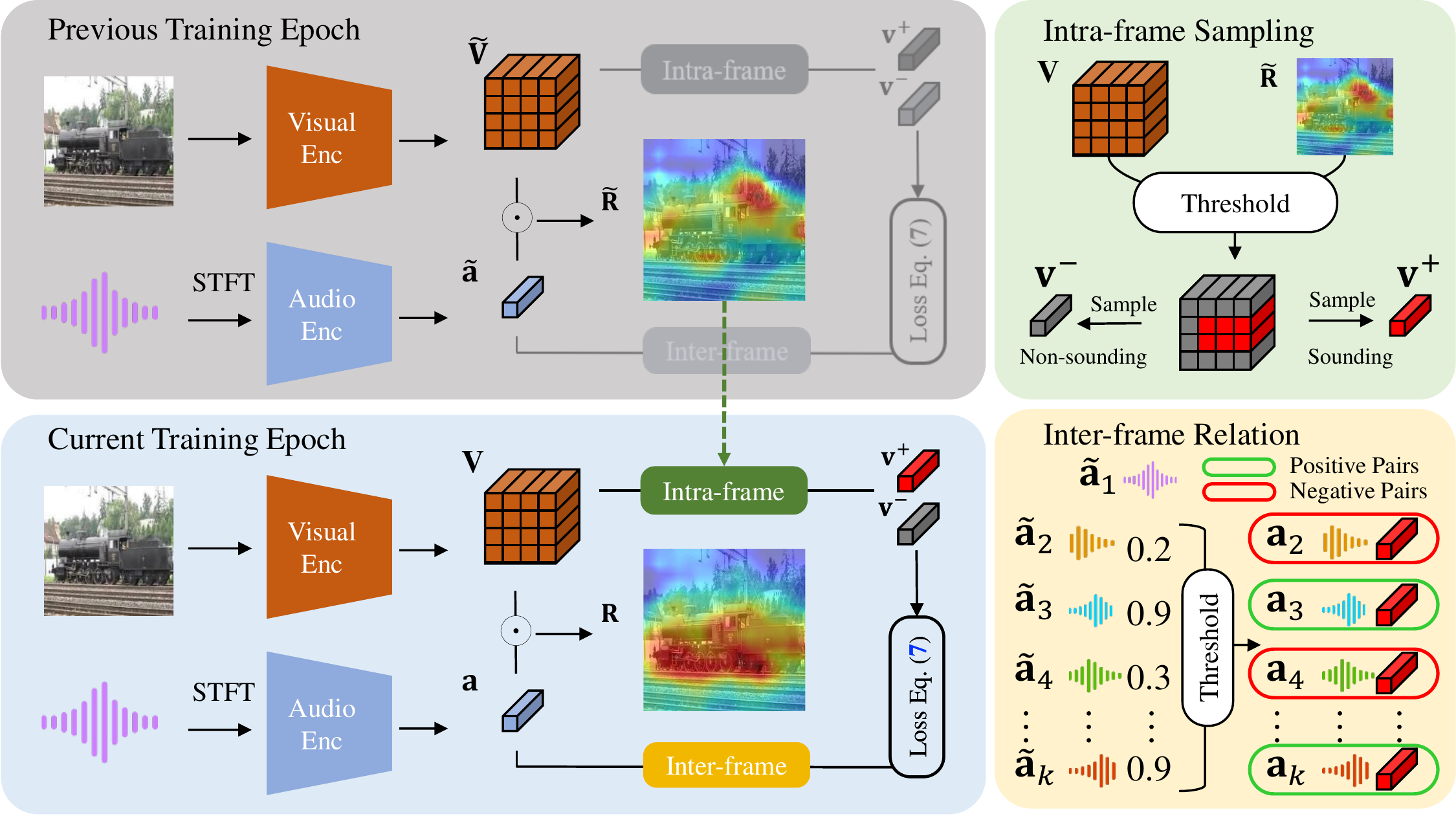}
 	\vspace{-3mm}
     \caption{\textbf{Algorithm overview.} Our framework consists of a visual feature extractor, an audio feature extractor, an intra-frame sampling module, and an inter-frame relation module. 
     \textit{(upper-left)} Sound localization $\Tilde{R}$ is obtained by computing the correlation between the visual and audio features.
     \textit{(bottom-left)} 
     Our iterative contrastive learning scheme uses the localization results predicted in the \emph{previous} training epoch as the \emph{pseudo}-labels for the current epoch.
     \textit{(upper-right)}
     The intra-frame sampling module uses the pseudo-labels to extract (non-)sounding regions for enhancing the efficacy of the contrastive learning.
     %
     \textit{(lower-right)} 
     The inter-frame relation module determines the correlation of images and audios sampled across videos by observing the relationship in the audio modality.
    }
    \vspace{\figmargin}
	\label{fig:method}
\end{figure*}

\Paragraph{Sound Source Localization in Visual Scenes.}
This task aims to find corresponding sounding regions in images from audio signals. 
We categorize methods addressing this task into three groups.
The first group of work~\cite{av_eccv20_objs_vids,av_cvpr18_lls,av_tpami20_lls} leverages the correspondence between audio and visual signals for supervision.
These methods assume that the audio and visual features extracted from the same video clip should be more similar than those extracted from different clips.
Some sound localization methods~\cite{av_cvpr18_lls,av_tpami20_lls} are formulated in a semi-supervised way to deal with limited annotated data.
The second line of work uses the class activation map (CAM)~\cite{cvpr16_CAM} to determine discriminative regions for categorical prediction.
Owens~\etal~\cite{av_eccv18_Owens} learn the audio and visual representations by the audio-visual correspondence and perform sound localization using the CAM model. 
Similarly, given the event labels, Qian \etal~\cite{av_eccv20_mms_loc} use the CAM model to identify sounding regions and corresponding audio clips.
As such, the sound and visual object in the same event can be associated.
Finally, some models~\cite{av_cvpr19_deep_cluster,av_arxiv_curricumlum_av_clutser} utilize audio-visual clusters to model audio-visual relationships.
These methods cluster different frequencies of an audio signal and visual patches in the images.
The centers of the audio and visual clusters extracted from the same video clip are associated during the training stage.

We note that existing sound localization approaches are limited in several aspects. 
These methods typically require additional information in other modalities (\eg optical flow~\cite{av_eccv20_objs_vids}), a pre-defined number of sound sources~\cite{av_cvpr19_deep_cluster,av_arxiv_curricumlum_av_clutser}, event labels in both audio and visual modalities~\cite{av_eccv20_mms_loc}, or single-source videos~\cite{av_nips20_loc}. 
In this work, we present a sound localization framework that does not rely on any additional annotation or assumption. 
Furthermore, the correlation between (non-)sounding objects and audio across pairs is jointly considered to further enhance sound localization.
%

\section{Methodology}
\label{sec:method}
\vspace{\secmargin}
\subsection{Sound Localization}
\vspace{\subsecmargin}
Our goal is to localize the source of the detected sound in the image.
Specifically, given the input image of size $W\times H\times 3$ and the detected audio, \ie sound, we aim to estimate the sounding region $\mathbf{S}$.
As shown in lower left panel of \figref{method}, the proposed sound localization model first extracts the corresponding visual representation $\mathbf{V}\in\mathbb{R}^{w\times h\times d}$ from the input image, and the audio feature representation $\mathbf{a}\in\mathbb{R}^{d}$ from the short-time Fourier-transformed~\cite{stft} audio.
We then use the attention mechanism to compute the response map $\mathbf{R}\in\mathbb{R}^{w\times h\times 1}$ followed by min-max normalization,
\vspace{\eqmargin}
\begin{equation}
\begin{aligned}
    \label{eq:normalization}
    \mathbf{R} &= \mathbf{V} * \mathbf{a},\\
    \mathbf{R} &= \frac{\mathbf{R} - \min(\mathbf{R})}{\max(\mathbf{R}) - \min(\mathbf{R})},
    \end{aligned}
    \vspace{\eqmargin}
\end{equation}
where the notation $*$ represents the pixel-wise inner-product operation. 
We then determine the potential sounding region by thresholding the response map $\mathbf{R}$:
\vspace{\eqmargin}
\begin{equation}
    \label{eq:cos}
    \mathbf{S} = \mathrm{idx}(\mathbf{R} > \delta_{v}),
    \vspace{\eqmargin}
\end{equation}
where $\delta_v\in[0,1]$ is a parameter for thresholding. 
The function $\mathrm{idx}(\cdot)$ returns the spatial indexes of the sampled patches that match the given condition. 
%
%

In the following, we will illustrate how the proposed method learns to localize sound via audio-visual representation learning.
The baseline audio-visual contrastive learning is first introduced.
It is used for initializing our model.
We then present our iterative training approach and finally discuss how we leverage the relationship given in the audio signals to facilitate the contrastive learning process.

\subsection{Audio-Visual Representation Learning}
\label{sec:av-learning}
\vspace{\subsecmargin}

\Paragraph{Contrastive Learning.} 
As audio-image pairs extracted from videos provide natural implication of the correlation between the two modalities, we use contrastive learning~\cite{infonce} to learn the audio-visual feature representations in an unsupervised manner.
The core idea is to maximize the correlation between the audio and visual representations extracted from the same video (\ie positive pairs) while minimizing the correlation between those from different videos (\ie negative pairs).
Specifically, during the training stage, our model extracts a set of audio features $\{\mathbf{a}_1,\cdots,\mathbf{a}_k\}$ and a set of visual representations $\{\mathbf{V}_1,\cdots,\mathbf{V}_k\}$ from the input batch consisting of $k$ image-audio pairs sampled from the same videos.
Then the model is optimized by the following training objective: 
\vspace{\eqmargin}
{\small
\begin{equation}
    \label{eq:nce_loss}
    \mathcal{L}_\mathrm{contrast} = -\frac{1}{k}\sum_{i=1}^k\Big[\log\frac{\exp(\phi(\mathbf{V}_i)\cdot\mathbf{a}_i/\tau)}{\sum_{j=1}^k\exp(\phi(\mathbf{V}_i)\cdot\mathbf{a}_j/\tau)}\Big],
    \vspace{\eqmargin}
\end{equation}}
where the term $\tau$ is a hyper-parameter controlling the temperature.
The notation $\phi$ represents the operations of $L2$ normalization on the feature dimension followed by average pooling on the spatial dimensions.

\Paragraph{Iterative Contrastive Learning.}
Since an image typically contains both sounding and non-sounding regions, the training loss in Eq.~\eqref{eq:nce_loss} is less effective as it takes the whole image into consideration at a time, which may associate non-sounding regions with the audio signals extracted from the same video.
Moreover, the annotations of the sounding objects are not available under the unsupervised setting.

To this end, we develop an iterative contrastive learning approach.
As illustrated in \figref{method}, starting from using conventional contrastive learning in Eq.~\ref{eq:nce_loss} for initialization, we take the sound localization results predicted in the \emph{previous} training epoch as the \emph{pseudo}-labels for \emph{current} training epoch.
Specifically, let $\mathbf{\Tilde{R}}_i= \mathbf{\Tilde{V}}_i * \mathbf{\Tilde{a}}_i$ denote the response map predicted from the model with parameters in the previous training epoch.
We randomly sample the visual features from patches, which show high responses on the map $\mathbf{\Tilde{R}}_i$ in the previous epoch, as the sounding feature $\mathbf{v}^+$ \ie
\vspace{\eqmargin}
\begin{equation}
\begin{aligned}
\label{eq:pos_idx}
\mathbf{x}^\mathrm{pos}_i &= \mathrm{idx}(\mathbf{\Tilde{R}}_i > \delta_{v}), \\
\mathbf{v}^{+}_i&= \phi(\mathrm{feats}({\mathbf{V}_i,\mathbf{x}^\mathrm{pos}_i})),\quad i=1\sim k,
\end{aligned}
\vspace{\eqmargin}
\end{equation}
%
where function $\mathrm{feats}(\cdot)$ returns a set of visual features for the given indexes. 
We replace the term $\phi(\mathbf{V}_i)$  in \eqnref{nce_loss} with the sounding feature $\mathbf{v}^+_i$.
In this way, the sounding regions are iteratively explored while non-sounding regions are gradually excluded.
In practice, we perform min-max normalization for $\mathbf{\Tilde{R}}_i$, same as \eqnref{normalization}, to prevent the threshold~$\delta_{v}$ too high to find confident sounding patches.

\Paragraph{Intra-Frame Sampling.}
We enhance the efficacy of the proposed contrastive learning by incorporating more negative pairs.
However, merely sampling more negative pairs by extracting audio and images from different videos is less effective as the model may easily determine the correlation.
Consequently, we propose to use the \emph{pseudo-non-sounding} regions predicted in the previous training epoch to form the negative pairs with the audio clips extracted from the same video.
We illustrate the process in \figref{teaser} (red line and red dotted circle) and \figref{method} (top right).
The correlation of these negative pairs is more challenging to determine as they are sampled from the \emph{same} video sequence, thus helping the sound localization model to learn more discriminative audio-visual representations.
We call such a strategy intra-frame sampling, which is formulated as follows:
\vspace{\eqmargin}
\begin{equation}
\begin{aligned}
\label{eq:neg_idx}
\mathbf{x}^\mathrm{neg}_i &= \mathrm{idx}(\mathbf{\Tilde{R}}_i < \delta_{v}), \\
\mathbf{v}^-_i&= \phi(\mathrm{feats}({\mathbf{V}_i, \mathbf{x}^\mathrm{neg}_i})),\quad i=1\sim k.
\end{aligned}
\vspace{\eqmargin}
\end{equation}

\Paragraph{Inter-Frame Relation.}
As the semantically similar contents may appear in different video sequences, contrastive learning can be further improved if it explores the correlation between images and audio signals from different videos.
An example is given in \figref{teaser} (black line and green dotted region).
Specifically, we leverage the relationship in the audio modality to determine the correlation of the image and audio clip sampled from different videos.
The relationship in the audio modality is estimated by using the audio representations $\mathbf{\Tilde{a}}$ computed in the \emph{previous} training epoch.
As shown in the bottom-right corner of \figref{method}, we determine the correlation $y_{ij} \in \{0,1\}$ between the $i$-th image and the $j$-th audio within the same mini-batch according to the audio representations, \ie
\vspace{\eqmargin}
\begin{equation}
\begin{aligned}
\label{eq:a_th}
y_{i,j}=\left\{\begin{array}{ll}
1, & \text { if } \left<\mathbf{\Tilde{a}}_{i}, \mathbf{\Tilde{a}}_{j}\right> \geq \delta_{a}, \\
0, & \text { otherwise,}\\ 
\end{array}\right. 
\; \forall i, j \in \{1, \ldots,k\},
\end{aligned}
\vspace{\eqmargin}
\end{equation}
where the term $\delta_{a} \in [0,1]$ is a thresholding parameter. 
Combining the proposed intra-frame sampling and inter-frame relation strategies, our training objective becomes
\vspace{\eqmargin}
{\small
\begin{equation}
\label{eq:full}
\begin{aligned}
&\mathcal{L}^\mathrm{iterative}_\mathrm{contrast}=\\
&-\frac{1}{k}\sum_{i=1}^k\Big[\log\frac{\sum_{j=1}^{k}y_{i,j}\exp(\textbf{v}_i^+\cdot\mathbf{a}_j/\tau)}{\sum_{j=1}^{k}\exp(\textbf{v}_i^-\cdot\mathbf{a}_j/\tau)+\exp(\textbf{v}_i^+\cdot\mathbf{a}_j/\tau)}\Big].
\end{aligned}
\vspace{\eqmargin}
\end{equation}
}
We train our sound localization model using Eq.~\eqref{eq:nce_loss} at the initialization stage, and then iteratively optimize the objective in Eq.~\eqref{eq:full} until the localization results converge.

\begin{table*}[t]
    \caption{\textbf{Quantitative results of sound localization.} 
    We evaluate all methods on the SoundNet-Flickr~\cite{av_cvpr18_lls,av_tpami20_lls} and MUSIC-Synthetic~\cite{av_nips20_loc} datasets with cIoU and AUC metrics. Following the evaluation protocol in~\cite{av_cvpr18_lls,av_tpami20_lls,av_nips20_loc}, we evaluate the cIoU@$0.5$ and cIoU@$0.3$ for SoundNet-Flickr  and MUSIC-Synthetic, respectively. 
    }
    \vspace{-3mm}
    \label{tab:stoa}
    \centering
    \normalsize
    \begin{tabular}{l cc cc cc}
        \toprule
        \multirow{2}{*}{Method} & \multicolumn{2}{c} {SoundNet-Flickr 10K }  & \multicolumn{2}{c} {SoundNet-Flickr 20K}& \multicolumn{2}{c} {MUSIC-Synthetic} \\
        \cline {2-7}
         & cIoU@0.5$\uparrow$ & AUC $\uparrow$ & cIoU@0.5 $\uparrow$ & AUC $\uparrow$ & cIoU@0.3 $\uparrow$ & AUC\\
        \midrule
        Random & $7.2$ & $30.7$ & $-$ & $-$ & $0.2$ & $9.6$\\
        Attention ~\cite{av_cvpr18_lls} & $42.1$ & $43.8$ & $45.3$ & $46.7$ & $6.9$ & $14.2$\\
        DMC ~\cite{av_cvpr19_deep_cluster} & $41.4$ & $45.0$ &$47.0$ & $47.5$& $6.6$ & $15.3$\\
        MSSL ~\cite{av_eccv20_mms_loc} & $51.2$ & $50.4$ & $53.8$ & $50.6$& $4.3$ & $12.1$\\
        DSOL ~\cite{av_nips20_loc} & $56.6$ & $51.5$ & $58.7$ & $52.9$  & $15.4$ & $17.0$\\
        \midrule
         Ours & $\mathbf{71.0}$ & $\mathbf{58.0}$ & $\mathbf{74.7}$ & $\mathbf{59.6}$& $\mathbf{25.1}$ & $\mathbf{21.9}$\\
        \bottomrule
    \end{tabular}
    \vspace{\tabmargin}
\end{table*}

\begin{figure}[t]
    \centering
	\includegraphics[width=0.8\linewidth]{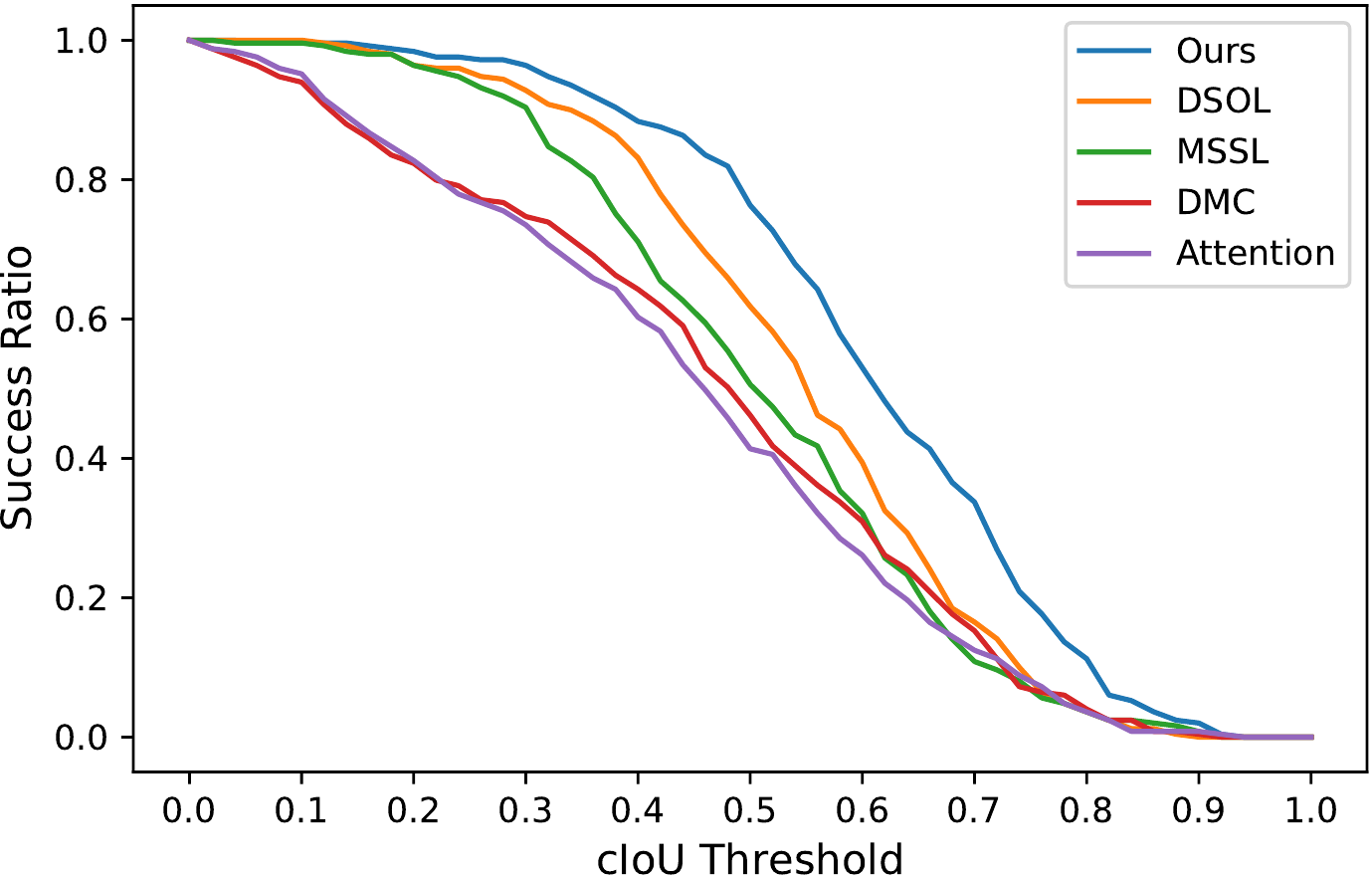}
 	\vspace{-2mm}
     \caption{\textbf{Success ratio under different cIoU thresholds.}
     Success ratio indicates the ratio of all instances whose cIoU scores are higher than  thresholds. Note that a larger area under the curve (AUC) indicates better performance.
     }
	\vspace{\figmargin}
	\vspace{-2mm}
	\label{fig:curve}
\end{figure}

\captionsetup[subfigure]{labelformat=empty} 
\begin{figure*}[t!]
\centering
    %
    \begin{subfigure}[b]{.16\linewidth}
    \centering
    \includegraphics[width=0.95\textwidth]{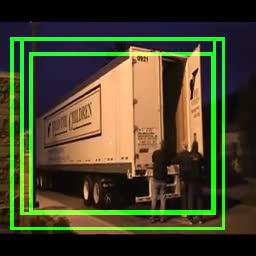}
    \end{subfigure}
    \begin{subfigure}[b]{.16\linewidth}
    \centering
    \includegraphics[width=0.95\textwidth]{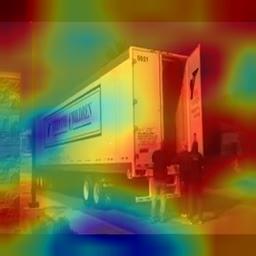}
    \end{subfigure}
    \begin{subfigure}[b]{.16\linewidth}
    \centering
    \includegraphics[width=0.95\textwidth]{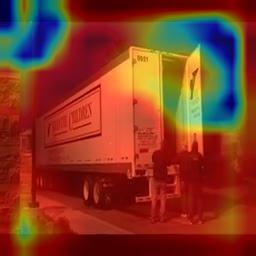}
    \end{subfigure}
    \begin{subfigure}[b]{.16\linewidth}
    \centering
    \includegraphics[width=0.95\textwidth]{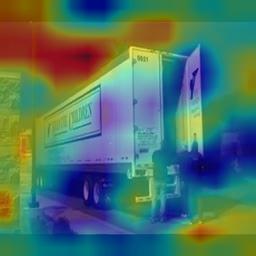}
    \end{subfigure}
    \begin{subfigure}[b]{.16\linewidth}
    \centering
    \includegraphics[width=0.95\textwidth]{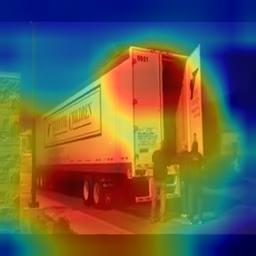}
    \end{subfigure}
    \begin{subfigure}[b]{.16\linewidth}
    \centering
    \includegraphics[width=0.95\textwidth]{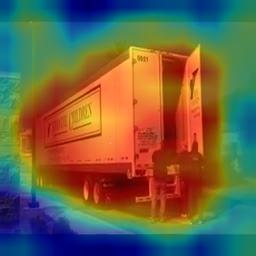}
    \end{subfigure}
    \\
    \begin{subfigure}[b]{.16\linewidth}
    \centering
    \includegraphics[width=0.95\textwidth]{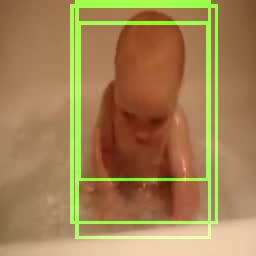}
    \end{subfigure}
    \begin{subfigure}[b]{.16\linewidth}
    \centering
    \includegraphics[width=0.95\textwidth]{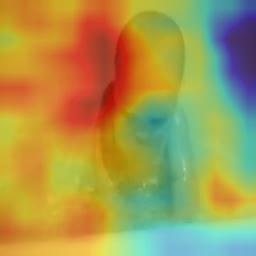}
    \end{subfigure}
    \begin{subfigure}[b]{.16\linewidth}
    \centering
    \includegraphics[width=0.95\textwidth]{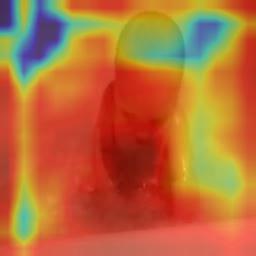}
    \end{subfigure}  
    \begin{subfigure}[b]{.16\linewidth}
    \centering
    \includegraphics[width=0.95\textwidth]{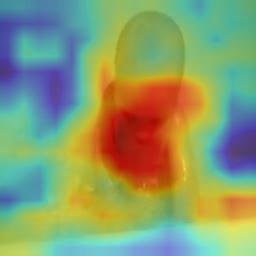}
    \end{subfigure}
    \begin{subfigure}[b]{.16\linewidth}
    \centering
    \includegraphics[width=0.95\textwidth]{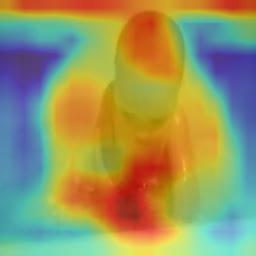}
    \end{subfigure}
    \begin{subfigure}[b]{.16\linewidth}
    \centering
    \includegraphics[width=0.95\textwidth]{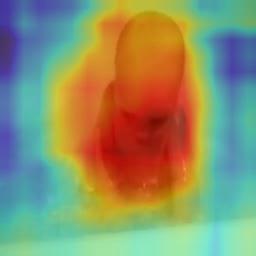}
    \end{subfigure}
    \\
    %
    \begin{subfigure}[b]{.16\linewidth}
    \centering
    \includegraphics[width=0.95\textwidth]{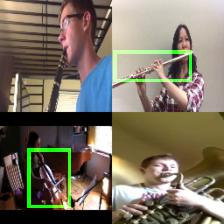}
    \end{subfigure}
    \begin{subfigure}[b]{.16\linewidth}
    \centering
    \includegraphics[width=0.95\textwidth]{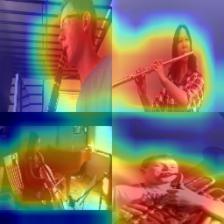}
    \end{subfigure}
    \begin{subfigure}[b]{.16\linewidth}
    \centering
    \includegraphics[width=0.95\textwidth]{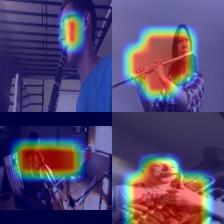}
    \end{subfigure}  
    \begin{subfigure}[b]{.16\linewidth}
    \centering
    \includegraphics[width=0.95\textwidth]{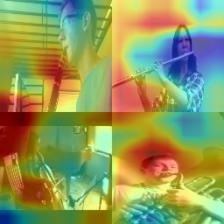}
    \end{subfigure}
    \begin{subfigure}[b]{.16\linewidth}
    \centering
    \includegraphics[width=0.95\textwidth]{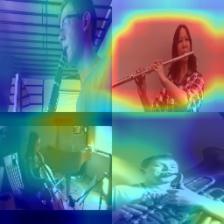}
    \end{subfigure}
    \begin{subfigure}[b]{.16\linewidth}
    \centering
    \includegraphics[width=0.95\textwidth]{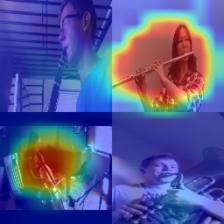}
    \end{subfigure}
    \begin{subfigure}[b]{.16\linewidth}
    \centering
    \includegraphics[width=0.95\textwidth]{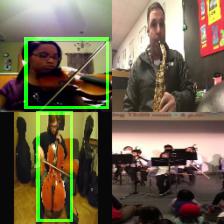}
    \caption{Input Image}
    \end{subfigure}
    \begin{subfigure}[b]{.16\linewidth}
    \centering
    \includegraphics[width=0.95\textwidth]{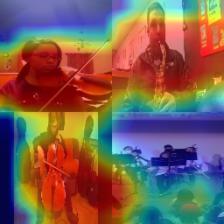}
    \caption{Attention~\cite{av_cvpr18_lls}}
    \end{subfigure}
    \begin{subfigure}[b]{.16\linewidth}
    \centering
    \includegraphics[width=0.95\textwidth]{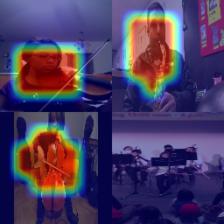}
    \caption{DMC~\cite{av_cvpr19_deep_cluster}}
    \end{subfigure}  
    \begin{subfigure}[b]{.16\linewidth}
    \centering
    \includegraphics[width=0.95\textwidth]{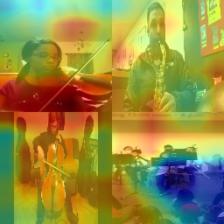}
    \caption{MSSL~\cite{av_eccv20_mms_loc}}
    \end{subfigure}
    \begin{subfigure}[b]{.16\linewidth}
    \centering
    \includegraphics[width=0.95\textwidth]{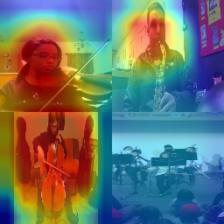}
    \caption{DSOL~\cite{av_nips20_loc}}
    \end{subfigure}
    \begin{subfigure}[b]{.16\linewidth}
    \centering
    \includegraphics[width=0.95\textwidth]{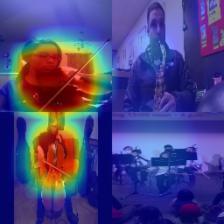}
    \caption{Ours}
    \end{subfigure}
  \caption{\textbf{Qualitative comparisons.} 
  We compare with state-of-the-art sound localization methods on the SoundNet-Flicker~\cite{av_cvpr18_lls,av_tpami20_lls}  (1st and 2nd rows) and MUSIC-Synthetic~\cite{av_nips20_loc} (3rd and 4th rows) datasets. 
  Sound localization is presented using heat maps, in which red regions indicate the estimated sound source.
  Note that the bounding boxes are the annotations of sounding regions from multiple annotators for the SoundNet-Flicker dataset. 
  }\label{fig:vis}
  \vspace{\figmargin}
\end{figure*}

\section{Experimental Results}
\label{sec:results}
\vspace{\secmargin}
\Paragraph{Datasets.} We evaluate all methods on two datasets:\begin{compactitem}[$\bullet$]
    \item\textbf{SoundNet-Flickr}~\cite{av_nips16_soundnet} dataset consists of more than two million video sequences.
We use a $5$-second audio clip and the central frame of the $5$-seconds corresponding video clip, to form an input pair for the proposed framework.
Note that we do not rely on any annotation (\eg bounding boxes) for model training.
In all experiments, we perform the training process with the subsets of the SoundNet-Flickr dataset constructed by Qian~\etal~\cite{av_eccv20_mms_loc} that contains $10$k and $20$k audio-visual pairs.
Following the protocol in~\cite{av_cvpr18_lls,av_eccv20_mms_loc,av_tpami20_lls}, we conduct the evaluation using the testing set of the SoundNet-Flickr dataset which consists of $250$ audio-visual pairs with bounding box annotations.
%
%
%
%
 \item{\textbf{MUSIC-Synthetic}}~\cite{av_nips20_loc} is a dataset consisting of synthetic audio-visual pairs.
 Each audio-visual pair is constructed by concatenating four music instrument frames and randomly selecting two out of four corresponding 1-second audios. 
 In other words, for each audio-visual pair, there are two instruments making sound while the other two are silent.
 We follow the protocol~\cite{av_nips20_loc} to train the models with all $25$k audio-visual pairs in the training set and conduct the evaluation on the testing set consisting of $455$ audio-visual pairs with bounding box annotations.

\end{compactitem}

\Paragraph{Implementation Details.} We implement the proposed method using Pytorch~\cite{pytorch}, and conduct the training and evaluation processes on a single NVIDIA GTX 1080 Ti GPU with $11$ GB memory.
We use the ResNet-18~\cite{resnet} architecture for both the visual and audio feature extractors.
Following the strategy in~\cite{av_eccv20_mms_loc,av_nips20_loc}, the visual feature extractor pre-trained on the ImageNet~\cite{ImageNet} dataset is employed.
As for the audio data pre-processing, the raw 5-seconds audio clips are re-sampled at $22.05$ kHz for the SoundNet-Flicker dataset (1-sceond clip at $16$kHz for the MUSIC-Synthetic dataset), and transformed into the log-mel spectrograms~(LMS).
%
%
Images are re-sized to the resolution of $256\times 256$ on SoundNet-Flicker and $224\times 224$ on MUSIC-Synthetic.
For fair comparisons, we adopt the same batch size of $96$ as in~\cite{av_eccv20_mms_loc,av_nips20_loc} for all the experiments. 
More implementation details are in the Supplementary.
The code and models will be made publicly available.

%

\Paragraph{Evaluation Metrics.}
Following previous work~\cite{av_cvpr19_deep_cluster,av_arxiv_curricumlum_av_clutser,av_eccv20_mms_loc,av_cvpr18_lls,av_tpami20_lls}, we adopt \textit{consensus intersection over union (cIoU)} and \textit{area under curve (AUC)} as the evaluation metrics.
Note that the ground-truth sounding region of an image is computed according to the overlapping of the bounding box labels annotated by different people.
The response map $\mathbf{R}$ in Eq.~\eqref{eq:normalization} is post-processed to serve as the sound localization results for evaluation.
Specifically, we first compute the response map $\mathbf{R}$ using Eq.~\eqref{eq:cos}.
Then we recover the resolution of the response map $\mathbf{R}$ from $w\times h$ to original image resolution $W \times H$ using bilinear up-sampling.

\captionsetup[subfigure]{labelformat=empty} 
\begin{figure*}[t!]
\captionsetup[subfigure]{justification=centering}
\centering
    \begin{subfigure}[b]{.16\linewidth}
    \centering
    \includegraphics[width=0.95\textwidth]{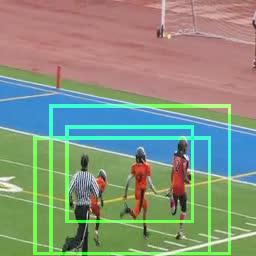}
    \end{subfigure}
    \begin{subfigure}[b]{.16\linewidth}
    \centering
    \includegraphics[width=0.95\textwidth]{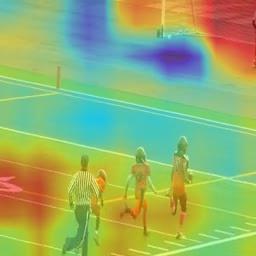}
    \end{subfigure}
    \begin{subfigure}[b]{.16\linewidth}
    \centering
    \includegraphics[width=0.95\textwidth]{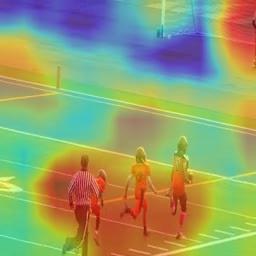}
    \end{subfigure}  
    \begin{subfigure}[b]{.16\linewidth}
    \centering
    \includegraphics[width=0.95\textwidth]{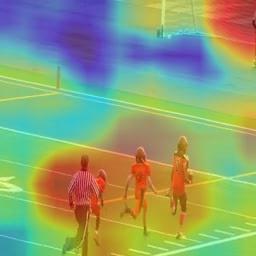}
    \end{subfigure}
    \begin{subfigure}[b]{.16\linewidth}
    \centering
    \includegraphics[width=0.95\textwidth]{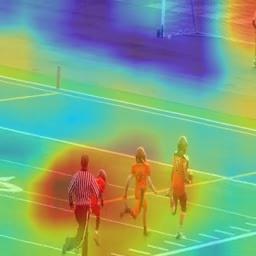}
    \end{subfigure}
    \begin{subfigure}[b]{.16\linewidth}
    \centering
    \includegraphics[width=0.95\textwidth]{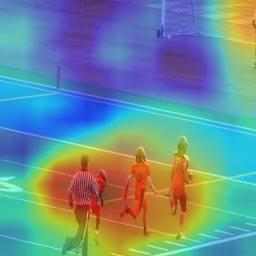}
    \end{subfigure}
    \\
    \begin{subfigure}[b]{.16\linewidth}
    \centering
    \includegraphics[width=0.95\textwidth]{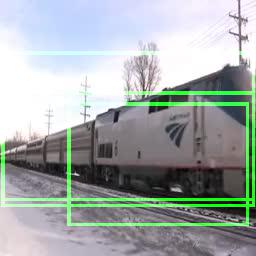}
    \end{subfigure}
    \begin{subfigure}[b]{.16\linewidth}
    \centering
    \includegraphics[width=0.95\textwidth]{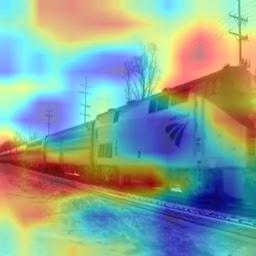}
    \end{subfigure}
    \begin{subfigure}[b]{.16\linewidth}
    \centering
    \includegraphics[width=0.95\textwidth]{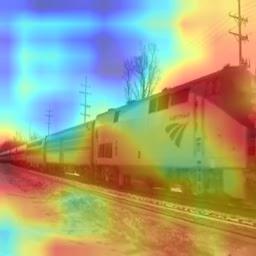}
    \end{subfigure}  
    \begin{subfigure}[b]{.16\linewidth}
    \centering
    \includegraphics[width=0.95\textwidth]{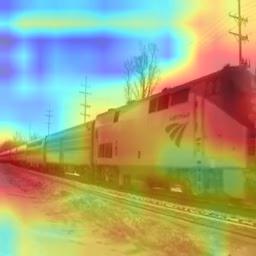}
    \end{subfigure}
    \begin{subfigure}[b]{.16\linewidth}
    \centering
    \includegraphics[width=0.95\textwidth]{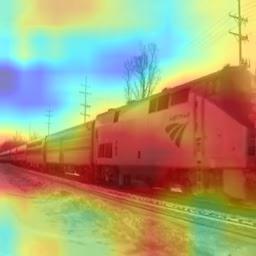}
    \end{subfigure}
    \begin{subfigure}[b]{.16\linewidth}
    \centering
    \includegraphics[width=0.95\textwidth]{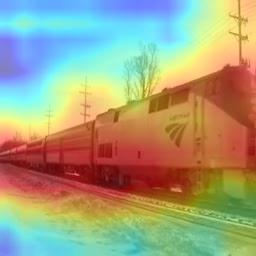}
    \end{subfigure}
    \\
    \begin{subfigure}[b]{.16\linewidth}
    \centering
    \includegraphics[width=0.95\textwidth]{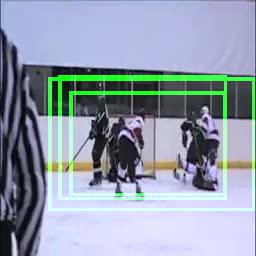}
    \caption{Input Image\\~}
    \end{subfigure}
    \begin{subfigure}[b]{.16\linewidth}
    \centering
    \includegraphics[width=0.95\textwidth]{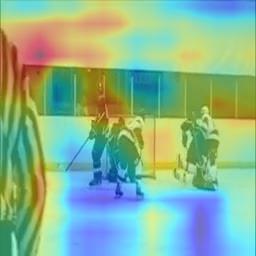}
    \caption{Conventional \\Contrastive Learning}
    \end{subfigure}
    \begin{subfigure}[b]{.16\linewidth}
    \centering
    \includegraphics[width=0.95\textwidth]{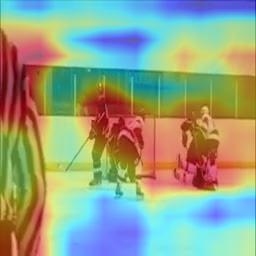}
    \caption{Iterative \\ Contrastive Learning}
    \end{subfigure}  
    \begin{subfigure}[b]{.16\linewidth}
    \centering
    \includegraphics[width=0.95\textwidth]{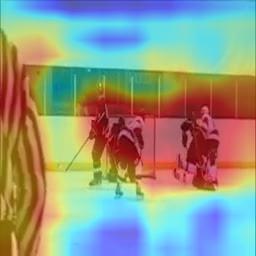}
    \caption{Iterative +\\ Intra-Frame}
    \end{subfigure}
    \begin{subfigure}[b]{.16\linewidth}
    \centering
    \includegraphics[width=0.95\textwidth]{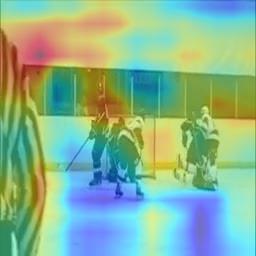}
    \caption{Iterative +\\ Inter-Frame}
    \end{subfigure}
    \begin{subfigure}[b]{.16\linewidth}
    \centering
    \includegraphics[width=0.95\textwidth]{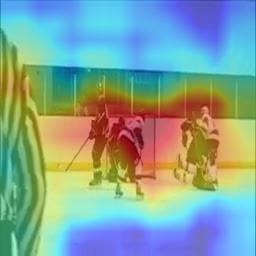}
    \caption{Full Model\\~}
    \end{subfigure}
  \caption{\textbf{Example localization results of using different design components in the proposed method on the SoundNet-Flicker~\cite{av_cvpr18_lls,av_tpami20_lls} dataset.} (\textit{from left to right}) We show the qualitative of conventional contrastive learning, iterative contrastive learning, iterative approach w/ intra-frame sampling, iterative approach w/ inter-frame relation, and our full model.
  }
  \label{fig:vis_abs}
  \vspace{\figmargin}
\end{figure*}

\Paragraph{Competing methods.}
We compare the proposed method to the following weakly- and unsupervised approaches:
\begin{compactitem}
\item\textbf{Attention}~\cite{av_cvpr18_lls,av_tpami20_lls} is trained using the audio-visual co-attention mechanism.

\item \textbf{DMC}~\cite{av_cvpr19_deep_cluster} is an unsupervised approach based on the usage of audio-visual clusters, and  requires \emph{a pre-defined number} of sound sources. We set the number of source to one suggested by ~\cite{av_cvpr19_deep_cluster,av_arxiv_curricumlum_av_clutser} for the SoundNet-Flick dataset and set to two for the MUSIC-Synthetic dataset.

\item \textbf{MSSL}~\cite{av_eccv20_mms_loc} reports the state-of-the-art performance on the sound localization task. It requires audio/visual event labels obtained from pre-trained classifiers and the CAM~\cite{cvpr16_CAM} predictions to find the sounding regions. 

\item \textbf{DSOL}~\cite{av_nips20_loc} is a two-stage approach requiring a large amount of single-source videos for the first stage to build up class-based visual dictionaries and train audio and visual encoders. %
For a fair comparison, we only train the network in the second stage.
We use pre-trained audio and visual encoders and use CAM~\cite{cvpr16_CAM} predictions to replace visual dictionaries.

\end{compactitem}

\subsection{Quantitative Results}
\vspace{-3.2mm}
%
\tabref{stoa} shows the quantitative comparisons on the SoundNet-Flickr and MUSIC-Synthetic datasets.
The proposed method performs favorably against the competing approaches on the sound localization task.
We note that different from the proposed method, the competing schemes require a pre-defined number of sounding sources~(\ie DMC) or audio/visual event labels~(\ie MSSL).
In contrast, the proposed method does not need any prior knowledge about the source number or data annotations.
Furthermore, our model trained with $10$k audio-visual pairs already outperforms MSSL and DSOL approaches which use more (\ie $20$k) audio-visual pairs during training.
In addition to the cIoU metric, the cIoU scores calculated with various thresholds are shown in \figref{curve}.
Our method reports favorable cIoU scores under all thresholds.
The consistent performance advantage suggests the effectiveness and efficacy of our iterative contrastive learning algorithm.

\begin{table}[t]
    \caption{
    \textbf{Ablation study.} (\textit{bottom}) We investigate the effect of using different design components in the proposed method. (\textit{top}) We show how we improve the MSSL approach by modifying the localziation method in~\eqnref{cos} and normalization strategy in~\eqnref{normalization}.}
    \vspace{-1mm}
    \label{tab:ablation}
    \centering
    \footnotesize
    \begin{tabular}{l cc}
        \toprule
        Method & cIoU@$0.5$ $\uparrow$ & AUC $\uparrow$ \\
        \midrule
        MSSL~\cite{av_eccv20_mms_loc} & $52.2$ & $49.6$ \\
        MSSL Stage I              & $42.2$ & $48.1$ \\
        MSSL Stage I w/o Labels   & $10.8$ & $30.2$ \\
        \midrule 
        MSSL Stage I w/~\eqnref{cos} & $47.4$ & $48.7$ \\
        MSSL Stage I w/o Labels w/~\eqnref{cos} & $47.0$ & $48.7$ \\
        \midrule
        MSSL Stage I w/~\eqnref{cos}~\eqnref{normalization} & $50.2$ & $49.0$ \\
        MSSL Stage I w/o Labels w/~\eqnref{cos}~\eqnref{normalization} & $46.6$ & $48.3$ \\
        \midrule
        Ours Initial                                        & $57.8$ & $52.1$ \\
        Ours Itr(\cmark) Intra(\xmark) Inter(\xmark)        & $64.2$ & $54.2$ \\
        Ours Itr(\cmark) Intra(\cmark) Inter(\xmark)        & $69.4$ & $56.9$ \\
        Ours Itr(\cmark) Intra(\xmark) Inter(\cmark)        & $67.1$ & $55.9$ \\
        \midrule
        Ours 10K & $\mathbf{71.0}$ & $\mathbf{58.0}$ \\
        \bottomrule
    \end{tabular}
\end{table}

\vspace{-0.2mm}
\subsection{Qualitative Evaluation}
\vspace{\subsecmargin}
We demonstrate the qualitative comparisons in \figref{vis}.
%
The localization results of the proposed method are more accurate compared to those of the competing approaches.
%
The example in the 3rd and 4th row is particularly challenging.
Since the multiple-sounding and non-sounding instruments appear in the same scene, it is difficult to localize exact-sounding objects.
MSSL and DSOL are both struggling with unrelated background. 
%
As for DMC, with the prior defined number of sounding source for the MUSIC-Synthetic dataset, it is more resistant to the unrelated background yet fail to identify the sounding instruments correctly.
%
%
Compared to these methods, the proposed framework can focus on the sounding objects with better accuracy, while trained without audio-visual event labels or any prior information.

%

\vspace{-0.2mm}
\subsection{Ablation Study}
\vspace{\subsecmargin}
\label{sec:ablation}

We conduct the ablation study to analyze the individual impact of each design component in the proposed method.
The results are presented in the fourth block of \tabref{ablation}, where \textbf{Itr} indicates the iterative contrastive training that uses the pseudo-sounding regions inferred from the previous epoch, \textbf{Intra} represents the usage of the pseudo-non-sounding regions, and \textbf{Inter} is the proposed inter-frame relation module.
We also demonstrate the qualitative comparisons in \figref{vis_abs}.
Particularly, the iterative strategy (\ie \textbf{Itr}) ensures the localization model focus only on the sounding region compared to the conventional contrastive learning approach (\ie \textbf{Initial}).
Both the quantitative and qualitative results confirm the efficacy of individual components designed in our approach.

\Paragraph{Comparison with MSSL.}
The proposed method shares similar backbone with the MSSL~\cite{av_eccv20_mms_loc} method.
%
%
Therefore, we also conduct the ablation study to show the impact of each modification we made, including replacing CAM with thresholding for sounding region  localization (Eq.~\eqref{eq:cos}), normalization (Eq.~\eqref{eq:normalization}), and conventional contrastive learaning (Eq.~\eqref{eq:nce_loss}).
The results are summarized in the first three blocks of \tabref{ablation}.
Since the MSSL method uses a two-stage model trained with audio-visual event labels, we study the case of removing the second stage (\textbf{Stage I}) and training without labels (\textbf{w/o Labels}).
As the results shown in the first block, training with the first stage and without labels both significantly degrade the performance of the MSSL method.
We show in the second and third block that using Eq.~\eqref{eq:cos} and Eq.~\eqref{eq:normalization} can greatly improve the performance.
Finally, we obtain our baseline (\textbf{Initial}) by applying Eq.~\eqref{eq:nce_loss} to the MSSL \textbf{Stage I} \textbf{w/o Labels} method with Eq.~\eqref{eq:cos} and Eq.~\eqref{eq:normalization}.
To conclude, \tabref{ablation} summarizes the impact of the proposed components and the transition from the original MSSL method to the proposed approach.

\begin{figure*}[t!]
    \centering
	\includegraphics[width=1\linewidth]{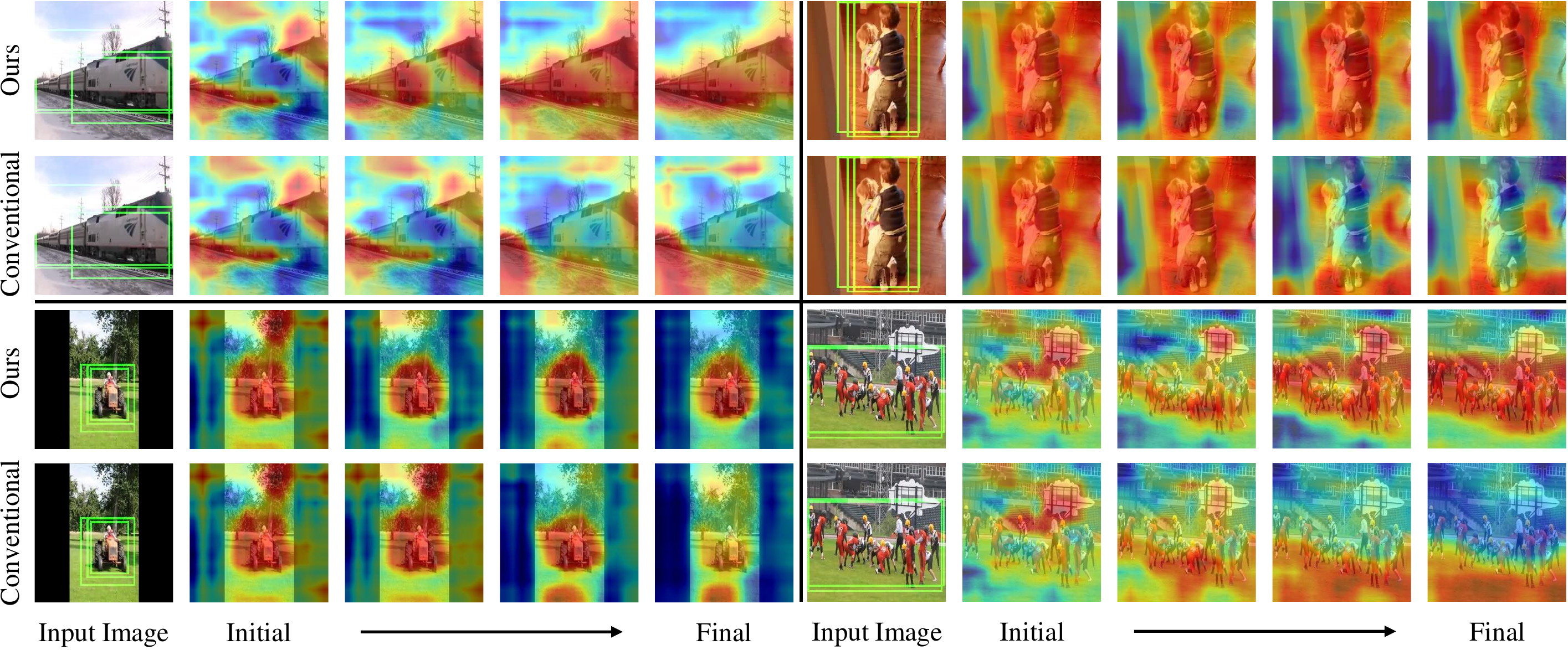}
 	\vspace{-6mm}
     \caption{\textbf{Example localization results at different training epochs on the SoundNet-Flicker dataset~\cite{av_cvpr18_lls,av_tpami20_lls}.} We present the sound localization results estimated by our method (1st and 3rd rows) and the conventional contrastive learning approach (2nd and 4th rows) at different (initial to final) training epochs.}
    \vspace{\figmargin}
    \vspace{2mm}
	 \label{fig:vis_itr}
\end{figure*}

\captionsetup[subfigure]{labelformat=empty}
\begin{figure*}[t]
\centering
  \begin{subfigure}[b]{.095\linewidth}
    \centering
    \caption{Reference}
    \vspace{-2mm}
    \includegraphics[width=1.0\textwidth]{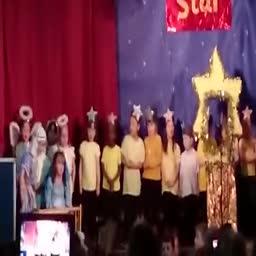}
  \end{subfigure}\hspace{-1mm}
  \begin{subfigure}[b]{0.095\linewidth}
    \centering
    \includegraphics[width=1.0\textwidth]{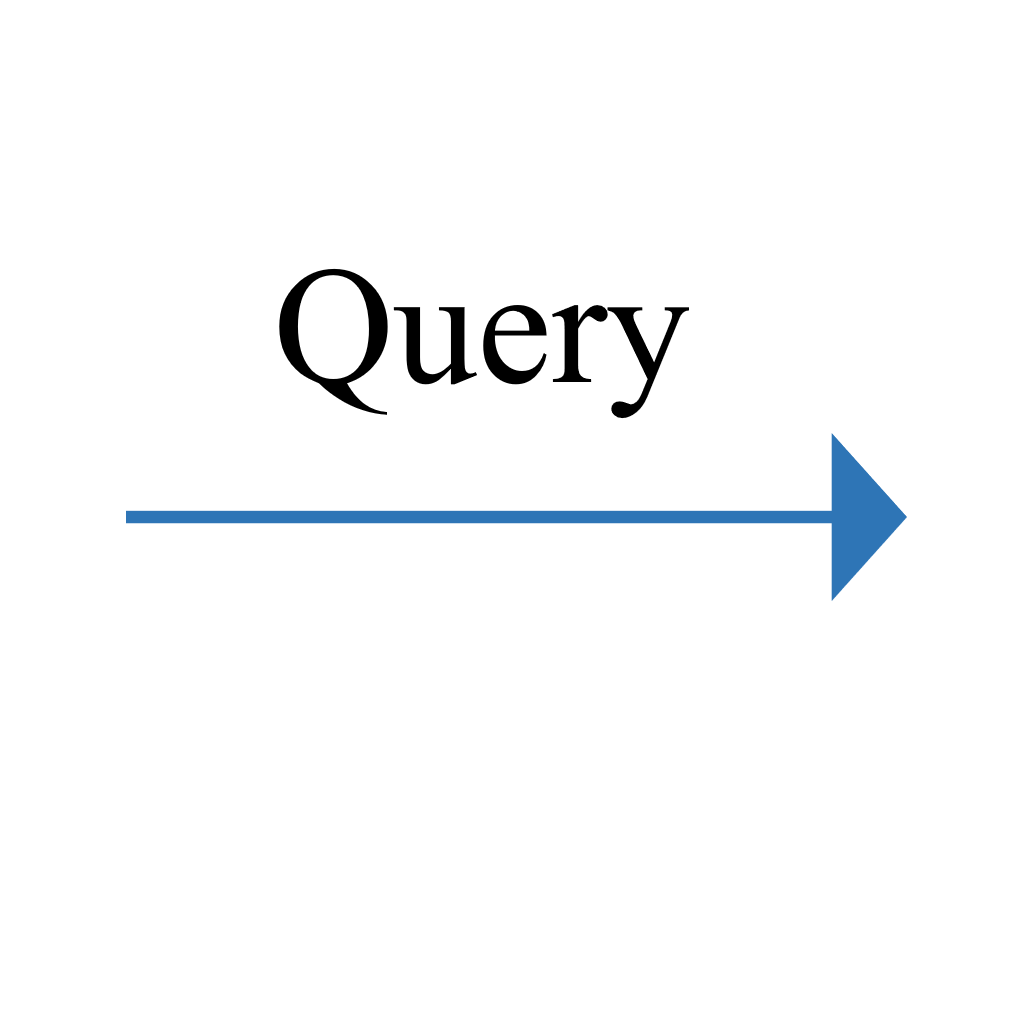}
  \end{subfigure} \hspace{-2mm} 
  \begin{subfigure}[b]{.095\linewidth}
    \centering
    \caption{Return \#1}
    \vspace{-2mm}
    \includegraphics[width=1.0\textwidth]{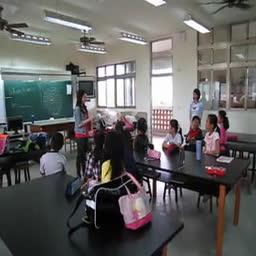}
  \end{subfigure}  
   \begin{subfigure}[b]{.095\linewidth}
    \centering
    \caption{Return \#2}
    \vspace{-2mm}
    \includegraphics[width=1.0\textwidth]{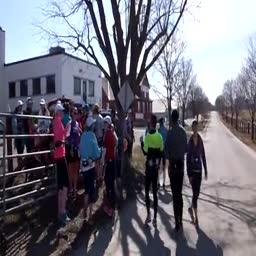}
  \end{subfigure} 
  \begin{subfigure}[b]{.095\linewidth}
    \centering
    \caption{Return \#3}
    \vspace{-2mm}
     \includegraphics[width=1.0\textwidth]{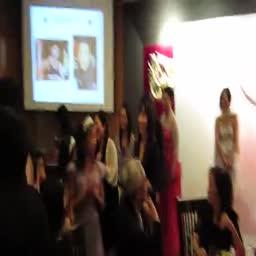}
  \end{subfigure} \hspace{5mm}
   \begin{subfigure}[b]{.095\linewidth}
    \centering
    \caption{Reference}
    \vspace{-2mm}
    \includegraphics[width=1.0\textwidth]{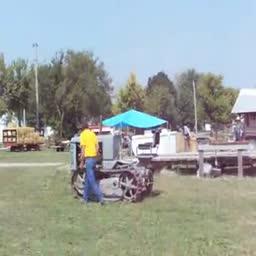}
  \end{subfigure}\hspace{-1mm} 
  \begin{subfigure}[b]{0.095\linewidth}
    \centering
    \includegraphics[width=1.0\textwidth]{figure/images/arrow.png}
  \end{subfigure}  \hspace{-2mm} 
\begin{subfigure}[b]{.095\linewidth}
    \centering
    \caption{Return \#1}
    \vspace{-2mm}
    \includegraphics[width=1.0\textwidth]{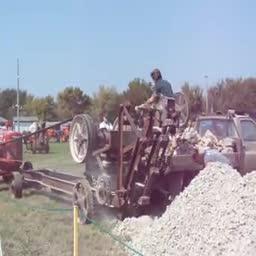}
\end{subfigure} 
\begin{subfigure}[b]{.095\linewidth}
    \centering
    \caption{Return \#2}
    \vspace{-2mm}
    \includegraphics[width=1.0\textwidth]{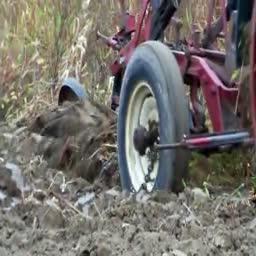}
\end{subfigure} 
\begin{subfigure}[b]{.095\linewidth}
    \centering
    \caption{Return \#3}
    \vspace{-2mm}
    \includegraphics[width=1.0\textwidth]{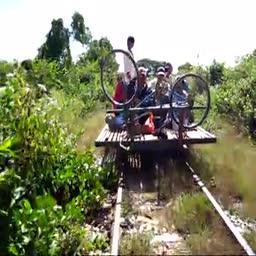}
\end{subfigure} 
\\
  \begin{subfigure}[b]{.095\linewidth}
    \centering
    \includegraphics[width=1.0\textwidth]{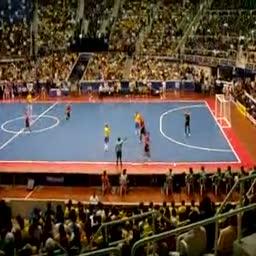}
  \end{subfigure}\hspace{-1mm}
  \begin{subfigure}[b]{0.095\linewidth}
    \centering
    \includegraphics[width=1.0\textwidth]{figure/images/arrow.png}
  \end{subfigure} \hspace{-2mm} 
  \begin{subfigure}[b]{.095\linewidth}
    \centering
    \includegraphics[width=1.0\textwidth]{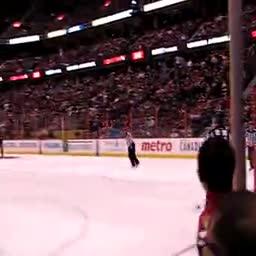}
  \end{subfigure}  
   \begin{subfigure}[b]{.095\linewidth}
    \centering
    \includegraphics[width=1.0\textwidth]{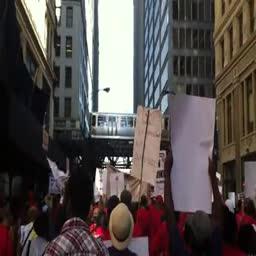}
  \end{subfigure} 
  \begin{subfigure}[b]{.095\linewidth}
    \centering
     \includegraphics[width=1.0\textwidth]{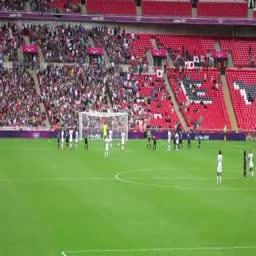}
  \end{subfigure} \hspace{5mm}
   \begin{subfigure}[b]{.095\linewidth}
    \centering
    \includegraphics[width=1.0\textwidth]{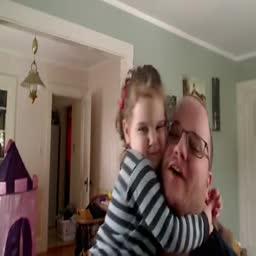}
  \end{subfigure}\hspace{-1mm} 
  \begin{subfigure}[b]{0.095\linewidth}
    \centering
    \includegraphics[width=1.0\textwidth]{figure/images/arrow.png}
  \end{subfigure}  \hspace{-2mm} 
\begin{subfigure}[b]{.095\linewidth}
    \centering
    \includegraphics[width=1.0\textwidth]{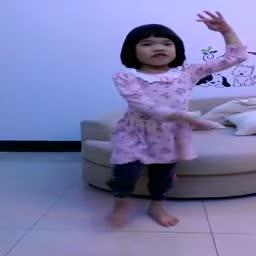}
\end{subfigure} 
\begin{subfigure}[b]{.095\linewidth}
    \centering
    \includegraphics[width=1.0\textwidth]{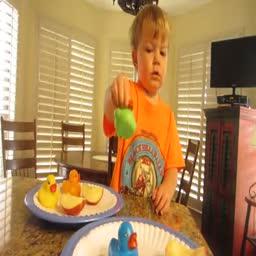}
\end{subfigure} 
\begin{subfigure}[b]{.095\linewidth}
    \centering
    \includegraphics[width=1.0\textwidth]{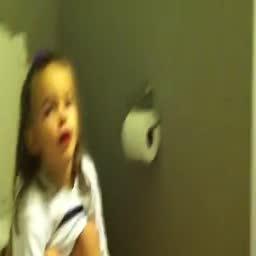}
\end{subfigure} 
  \\%
  \begin{subfigure}[b]{.095\linewidth}
    \centering
    \includegraphics[width=1.0\textwidth]{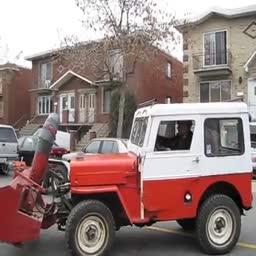}
  \end{subfigure}\hspace{-1mm}
  \begin{subfigure}[b]{0.095\linewidth}
    \centering
    \includegraphics[width=1.0\textwidth]{figure/images/arrow.png}
  \end{subfigure} \hspace{-2mm} 
  \begin{subfigure}[b]{.095\linewidth}
    \centering
    \includegraphics[width=1.0\textwidth]{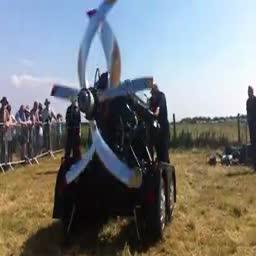}
  \end{subfigure}  
   \begin{subfigure}[b]{.095\linewidth}
    \centering
    \includegraphics[width=1.0\textwidth]{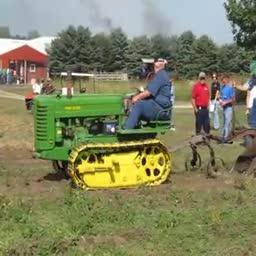}
  \end{subfigure} 
  \begin{subfigure}[b]{.095\linewidth}
    \centering
     \includegraphics[width=1.0\textwidth]{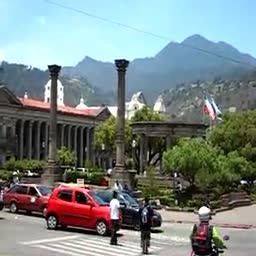}
  \end{subfigure} \hspace{5mm}
   \begin{subfigure}[b]{.095\linewidth}
    \centering
    \includegraphics[width=1.0\textwidth]{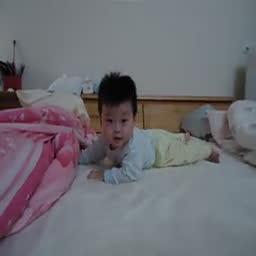}
  \end{subfigure}\hspace{-1mm} 
  \begin{subfigure}[b]{0.095\linewidth}
    \centering
    \includegraphics[width=1.0\textwidth]{figure/images/arrow.png}
  \end{subfigure}  \hspace{-2mm} 
\begin{subfigure}[b]{.095\linewidth}
    \centering
    \includegraphics[width=1.0\textwidth]{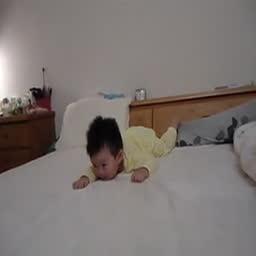}
\end{subfigure} 
\begin{subfigure}[b]{.095\linewidth}
    \centering
    \includegraphics[width=1.0\textwidth]{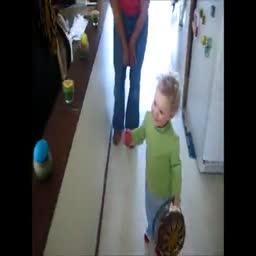}
\end{subfigure} 
\begin{subfigure}[b]{.095\linewidth}
    \centering
    \includegraphics[width=1.0\textwidth]{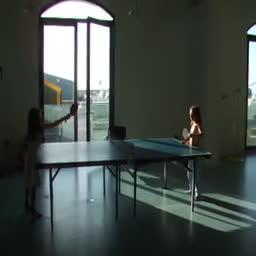}
\end{subfigure} 
  \\
  \vspace{-2mm}
  \caption{\textbf{Retrieval results from audio signals on the SoundNet-Flicker~\cite{av_cvpr18_lls,av_tpami20_lls}.} We use the sounds of the reference images as the \emph{queries} to retrieve the top-$3$ related audio clips and show the corresponding images.
  The results verify our idea that the relationships in the audio modality can help the association between images and audios extracted across videos.
  %
  }\label{fig:vis_inter}
  \vspace{\figmargin}
\end{figure*}

\Paragraph{Localization results in various epochs.}
Since the proposed iterative method is based on the strategy where the localization results predicted in the previous training epoch serve as the pseudo-label, the iterative localization results are crucial.
Therefore, we visualize the localization results at different epochs. 
%
%
As shown in~\figref{vis_itr}, the localization results gradually focus on the sounding regions.
The results validate the efficacy of the proposed iterative procedure that takes localization results from the previous epoch as training guidance for the current epoch.

\Paragraph{Relationships in audio modality.}
The proposed inter-frame relation illustrated in \secref{av-learning} is based on the assumption that the relationships in the audio modality can be the guidance of the contrasting learning.
To verify the assumption, we visualize the retrieval results in the audio modality in \figref{vis_inter}.
Specifically, given a reference audio-visual pair, we retrieve the top three audio-visual pairs according to the distances between audio features.
We present the images of the reference and retrieved visual-audio pairs in~\figref{vis_inter}.
%
%
As the reference and retrieved images share semantically similar contents, we validate the intuition behind the proposed inter-frame relation design.
%
%
%
%
%

\vspace{-2mm}
\section{Conclusions}
\vspace{\secmargin}
\label{sec:conclusions}
In this paper, we present a novel unsupervised sound localization framework that does not require any prior assumption or data annotation.
We propose two modules to provide pseudo positive and negative training pairs based on an iterative contrastive learning pipeline.
The \textit{intra-frame sampling} leverages the localization results estimated in the previous epoch as pseudo-labels.
The \textit{inter-frame relation} contributes  to training pairs across different videos by exploiting the relationships in the audio modality with the audio features learned from the previous epoch.
%
%
%
%
%
%
Extensive experimental results show that our approach performs favorably against the state-of-the-art weakly-supervised and unsupervised algorithms.

\clearpage
{\small
\bibliographystyle{ieee_fullname} 
\bibliography{egbib}

\begin{thebibliography}{10}\itemsep=-1pt

\bibitem{av_eccv20_objs_vids}
Triantafyllos Afouras, Andrew Owens, Joon~Son Chung, and Andrew Zisserman.
\newblock Self-supervised learning of audio-visual objects from video.
\newblock In {\em ECCV}, 2020.

\bibitem{avt_nips20_VersatileNet}
Jean-Baptiste Alayrac, Adri{\`a} Recasens, Rosalia Schneider, Relja
  Arandjelovi{\'c}, Jason Ramapuram, Jeffrey De~Fauw, Lucas Smaira, Sander
  Dieleman, and Andrew Zisserman.
\newblock Self-supervised multimodal versatile networks.
\newblock In {\em NeurIPS}, 2020.

\bibitem{av_nips20_xdc}
Humam Alwassel, Dhruv Mahajan, Lorenzo Torresani, Bernard Ghanem, and Du Tran.
\newblock Self-supervised learning by cross-modal audio-video clustering.
\newblock In {\em NeurIPS}, 2020.

\bibitem{av_iccv17_look}
Relja Arandjelovic and Andrew Zisserman.
\newblock Look, listen and learn.
\newblock In {\em ICCV}, 2017.

\bibitem{av_eccv18_obj_that_sound}
Relja Arandjelovi{\'c} and Andrew Zisserman.
\newblock Objects that sound.
\newblock In {\em ECCV}, 2018.

\bibitem{av_nips20_CrossLabelling}
Yuki~M Asano, Mandela Patrick, Christian Rupprecht, and Andrea Vedaldi.
\newblock Labelling unlabelled videos from scratch with multi-modal
  self-supervision.
\newblock In {\em NeurIPS}, 2020.

\bibitem{av_nips16_soundnet}
Yusuf Aytar, Carl Vondrick, and Antonio Torralba.
\newblock Soundnet: Learning sound representations from unlabeled video.
\newblock In {\em NeurIPS}, 2016.

\bibitem{ImageNet}
J. Deng, W. Dong, R. Socher, L.-J. Li, K. Li, and L. Fei-Fei.
\newblock {ImageNet: A Large-Scale Hierarchical Image Database}.
\newblock In {\em CVPR}, 2009.

\bibitem{av_cvpr20_sep-gesture}
Chuang Gan, Deng Huang, Hang Zhao, Joshua~B Tenenbaum, and Antonio Torralba.
\newblock Music gesture for visual sound separation.
\newblock In {\em CVPR}, 2020.

\bibitem{av_eccv18_sep}
Ruohan Gao, Rogerio Feris, and Kristen Grauman.
\newblock Learning to separate object sounds by watching unlabeled video.
\newblock In {\em ECCV}, 2018.

\bibitem{25d}
Ruohan Gao and Kristen Grauman.
\newblock 2.5d-visual-sound.
\newblock In {\em CVPR}, 2019.

\bibitem{co_sep_iccv19}
Ruohan Gao and Kristen Grauman.
\newblock Co-separating sounds of visual objects.
\newblock In {\em ICCV}, 2019.

\bibitem{av_cvpr21_gao2021VisualVoice}
Ruohan Gao and Kristen Grauman.
\newblock Visualvoice: Audio-visual speech separation with cross-modal
  consistency.
\newblock In {\em CVPR}, 2021.

\bibitem{av_cvpr20_PreviewAudio}
Ruohan Gao, Tae-Hyun Oh, Kristen Grauman, and Lorenzo Torresani.
\newblock Listen to look: Action recognition by previewing audio.
\newblock In {\em CVPR}, 2020.

\bibitem{stft}
D. {Griffin} and {Jae Lim}.
\newblock Signal estimation from modified short-time fourier transform.
\newblock In {\em ICASSP}, 1983.

\bibitem{resnet}
Kaiming He, Xiangyu Zhang, Shaoqing Ren, and Jian Sun.
\newblock Deep residual learning for image recognition.
\newblock In {\em CVPR}, 2016.

\bibitem{av_cvpr19_deep_cluster}
Di Hu, Feiping Nie, and Xuelong Li.
\newblock Deep multimodal clustering for unsupervised audiovisual learning.
\newblock In {\em CVPR}, 2019.

\bibitem{av_nips20_loc}
Di Hu, Rui Qian, Minyue Jiang, Xiao Tan, Shilei Wen, Errui Ding, Weiyao Lin,
  and Dejing Dou.
\newblock Discriminative sounding objects localization via self-supervised
  audiovisual matching.
\newblock In {\em NeurIPS}, 2020.

\bibitem{av_arxiv_curricumlum_av_clutser}
Di Hu, Zheng Wang, Haoyi Xiong, Dong Wang, Feiping Nie, and Dejing Dou.
\newblock Curriculum audiovisual learning.
\newblock {\em arXiv Preprint}, 2020.

\bibitem{av_nips18_coop}
Bruno Korbar, Du Tran, and Lorenzo Torresani.
\newblock Cooperative learning of audio and video models from self-supervised
  synchronization.
\newblock In {\em NeurIPS}, 2018.

\bibitem{av_iclr21_lee2021crossattentional}
Jun-Tae Lee, Mihir Jain, Hyoungwoo Park, and Sungrack Yun.
\newblock Cross-attentional audio-visual fusion for weakly-supervised action
  localization.
\newblock In {\em ICLR}, 2021.

\bibitem{AVSDN}
Yan-Bo Lin, Yu-Jhe Li, and Yu-Chiang~Frank Wang.
\newblock Dual-modality seq2seq network for audio-visual event localization.
\newblock In {\em ICASSP}, 2019.

\bibitem{my_accv20_av-trans}
Yan-Bo Lin and Yu-Chiang~Frank Wang.
\newblock Audiovisual transformer with instance attention for audio-visual
  event localization.
\newblock In {\em ACCV}, 2020.

\bibitem{my_aaai21_avconsistency}
Yan-Bo Lin and Yu-Chiang~Frank Wang.
\newblock Exploiting audio-visual consistency with partial supervision for
  spatial audio generation.
\newblock In {\em AAAI}, 2021.

\bibitem{lu2019self}
Yu-Ding Lu, Hsin-Ying Lee, Hung-Yu Tseng, and Ming-Hsuan Yang.
\newblock Self-supervised audio spatialization with correspondence classifier.
\newblock In {\em ICIP}. IEEE, 2019.

\bibitem{av_iclr21_activeContrastive}
Shuang Ma, Zhaoyang Zeng, Daniel McDuff, and Yale Song.
\newblock Active contrastive learning of audio-visual video representations.
\newblock In {\em ICLR}, 2021.

\bibitem{av_nips20_spatial_alignment}
Pedro Morgado, Yi Li, and Nuno Vasconcelos.
\newblock Learning representations from audio-visual spatial alignment.
\newblock In {\em NeurIPS}, 2020.

\bibitem{av_cvpr21_RAVID}
Pedro Morgado, Ishan Misra, and Nuno Vasconcelos.
\newblock Robust audio-visual instance discrimination.
\newblock In {\em CVPR}, 2021.

\bibitem{360gen}
Pedro Morgado, Nuno Nvasconcelos, Timothy Langlois, and Oliver Wang.
\newblock Self-supervised generation of spatial audio for 360 video.
\newblock In {\em NeurIPS}, 2018.

\bibitem{av_cvpr21_agreementAVID}
Pedro Morgado, Nuno Vasconcelos, and Ishan Misra.
\newblock Audio-visual instance discrimination with cross-modal agreement.
\newblock In {\em CVPR}, 2021.

\bibitem{infonce}
Aaron van~den Oord, Yazhe Li, and Oriol Vinyals.
\newblock Representation learning with contrastive predictive coding.
\newblock {\em arXiv Preprint}, 2018.

\bibitem{av_eccv18_Owens}
Andrew Owens and Alexei~A. Efros.
\newblock Audio-visual scene analysis with self-supervised multisensory
  features.
\newblock In {\em ECCV}, 2018.

\bibitem{av_eccv16_abSound}
Andrew Owens, Jiajun Wu, Josh~H McDermott, William~T Freeman, and Antonio
  Torralba.
\newblock Ambient sound provides supervision for visual learning.
\newblock In {\em ECCV}, 2016.

\bibitem{pytorch}
Adam Paszke, Sam Gross, Francisco Massa, Adam Lerer, James Bradbury, Gregory
  Chanan, Trevor Killeen, Zeming Lin, Natalia Gimelshein, Luca Antiga, Alban
  Desmaison, Andreas Kopf, Edward Yang, Zachary DeVito, Martin Raison, Alykhan
  Tejani, Sasank Chilamkurthy, Benoit Steiner, Lu Fang, Junjie Bai, and Soumith
  Chintala.
\newblock Pytorch: An imperative style, high-performance deep learning library.
\newblock In {\em NeurIPS}. 2019.

\bibitem{av_eccv20_mms_loc}
Rui Qian, Di Hu, Heinrich Dinkel, Mengyue Wu, Ning Xu, and Weiyao Lin.
\newblock Multiple sound sources localization from coarse to fine.
\newblock In {\em ECCV}, 2020.

\bibitem{av_cvpr18_lls}
Arda Senocak, Tae-Hyun Oh, Junsik Kim, Ming-Hsuan Yang, and In~So Kweon.
\newblock Learning to localize sound source in visual scenes.
\newblock In {\em CVPR}, 2018.

\bibitem{av_tpami20_lls}
Arda Senocak, Tae-Hyun Oh, Junsik Kim, Ming-Hsuan Yang, and In~So Kweon.
\newblock Learning to localize sound sources in visual scenes: Analysis and
  applications.
\newblock {\em TPAMI}, 2019.

\bibitem{av_cvpr21_co_learning}
Yapeng Tian, Di Hu, and Chenliang Xu.
\newblock Cyclic co-learning of sounding object visual grounding and sound
  separation.
\newblock In {\em CVPR}, 2021.

\bibitem{av_eccv20_avvp}
Yapeng Tian, Dingzeyu Li, and Chenliang Xu.
\newblock Unified multisensory perception: Weakly-supervised audio-visual video
  parsing.
\newblock In {\em ECCV}, 2020.

\bibitem{eccv18_avel}
Yapeng Tian, Jing Shi, Bochen Li, Zhiyao Duan, and Chenliang Xu.
\newblock Audio-visual event localization in unconstrained videos.
\newblock In {\em ECCV}, 2018.

\bibitem{av_iclr21_AudioScope}
Efthymios Tzinis, Scott Wisdom, Aren Jansen, Shawn Hershey, Tal Remez, Dan
  Ellis, and John~R. Hershey.
\newblock Into the wild with audioscope: Unsupervised audio-visual separation
  of on-screen sounds.
\newblock In {\em ICLR}, 2021.

\bibitem{av_cvpr21_av_parsing}
Yu Wu and Yi Yang.
\newblock Exploring heterogeneous clues for weakly-supervised audio-visual
  video parsing.
\newblock In {\em CVPR}, 2021.

\bibitem{av_iccv19_DAM}
Yu Wu, Linchao Zhu, Yan Yan, and Yi Yang.
\newblock Dual attention matching for audio-visual event localization.
\newblock In {\em ICCV}, 2019.

\bibitem{av_iccv19_mpnet}
Xudong Xu, Bo Dai, and Dahua Lin.
\newblock Recursive visual sound separation using minus-plus net.
\newblock In {\em ICCV}, 2019.

\bibitem{av_cvpr21_withoutBinaural}
Xudong Xu, Hang Zhou, Ziwei Liu, Bo Dai, Xiaogang Wang, and Dahua Lin.
\newblock Visually informed binaural audio generation without binaural audios.
\newblock In {\em CVPR}, 2021.

\bibitem{av_cvpr20_tell_left_right}
Karren Yang, Bryan Russell, and Justin Salamon.
\newblock Telling left from right: Learning spatial correspondence of sight and
  sound.
\newblock In {\em CVPR}, 2020.

\bibitem{sofm_iccv19}
Hang Zhao, Chuang Gan, Wei-Chiu Ma, and Antonio Torralba.
\newblock The sound of motions.
\newblock In {\em ICCV}, 2019.

\bibitem{pix}
Hang Zhao, Chuang Gan, Andrew Rouditchenko, Carl Vondrick, Josh McDermott, and
  Antonio Torralba.
\newblock The sound of pixels.
\newblock In {\em ECCV}, 2018.

\bibitem{cvpr16_CAM}
Bolei Zhou, Aditya Khosla, Agata Lapedriza, Aude Oliva, and Antonio Torralba.
\newblock Learning deep features for discriminative localization.
\newblock In {\em CVPR}, 2016.

\bibitem{av_eccv20_sep-stereo}
Hang Zhou, Xudong Xu, Dahua Lin, Xiaogang Wang, and Ziwei Liu.
\newblock Sep-stereo: Visually guided stereophonic audio generation by
  associating source separation.
\newblock In {\em ECCV}, 2020.

\end{thebibliography}
}

\end{document}